\documentclass[journal,12pt,onecolumn,draftclsnofoot,]{IEEEtran}

\ifCLASSINFOpdf
\else
   \usepackage[dvips]{graphicx}
\fi
\usepackage{url}
\usepackage{graphicx}

\usepackage{epsfig}
\usepackage{calc}
\usepackage{amssymb}
\usepackage{amstext}
\usepackage{amsmath}
\usepackage{amsthm}
\usepackage{multicol}
\usepackage{pslatex}
\usepackage{caption}
\usepackage{subcaption}
\usepackage{cite}
\usepackage{epsfig}
\usepackage{amsbsy} 
\usepackage{multirow}
\usepackage{multicol}
\usepackage{bm}
\usepackage{hhline}
\usepackage{algorithm}
\usepackage{algorithmic}
\usepackage{amsmath,amsfonts,amssymb}
\usepackage{graphicx}
\usepackage{verbatim}
\usepackage{enumitem}

\newtheorem{defn}{Definition}

\begin{document}

\title{Learning optimally separated  class-specific subspace  representations  using  convolutional autoencoder }

\author{Krishan Sharma, Shikha Gupta, and Renu Rameshan
		\thanks{Krishan Sharma, was with 	Indian Institute of Technology Mandi, India. He is now with the Vehant Technologies Pvt. Ltd., Noida, India, (e-mail: krishans@vehant.com).
	}
		\thanks{Shikha Gupta, was with 	Indian Institute of Technology Mandi India. He is now with the Vehant Technologies Pvt. Ltd. Noida, India, (e-mail: shikhag@vehant.com).
}
	\thanks{Renu Rameshan is with School of Computing and Electrical Engineering,
		Indian Institute of Technology Mandi, India  (e-mail: renumr@iitmandi.ac.in).}

}

%

\maketitle

\begin{abstract}
In this work, we propose   a  novel convolutional  autoencoder based architecture to generate  subspace specific  feature representations that  are   best suited for classification task.  The class-specific data is assumed to   lie in  low dimensional linear subspaces, which could be noisy  and not well separated, i.e., subspace distance (principal angle) between two classes is very low. The proposed  network   uses a  novel class-specific self expressiveness (CSSE) layer sandwiched  between encoder and decoder networks  to generate  class-wise subspace representations which are well separated. The CSSE layer along with encoder/ decoder are trained in such a way that data still lies in subspaces in the feature space with minimum principal angle much higher than that of the input space. To demonstrate the effectiveness of the  proposed approach, several experiments have been carried out  on  state-of-the-art machine learning datasets and a significant improvement  in classification performance is observed over existing subspace based transformation learning methods. 
\end{abstract}


\IEEEpeerreviewmaketitle

\section{Introduction}
\label{sec:intro}
In most of the machine learning and computer vision applications,  high dimensional  data usually resides in some linear or non-linear geometric structures of a comparably small intrinsic dimension. Subspaces and manifolds are  well known examples of  such linear and non-linear  structures, respectively. Presence  of such structures in  high dimensional data is established in ~\cite{basri2003lambertian,hastie1998metrics, murase1995visual,tomasi1992shape}. Basri and Jacob~\cite{basri2003lambertian} established that face images of a particular subject in varying lighting conditions  can be well approximated by a  nine  dimensional linear subspace of a  high dimensional ambient space under the assumption of Lambertian reflectance. On  similar lines, face images from  multiple subjects can be assumed to lie in a union of low dimensional linear subspaces. Similar studies have been carried out for images of handwritten  digit datasets~\cite{hastie1998metrics},   object images under different orientations~\cite{murase1995visual} and trajectories of moving objects~\cite{tomasi1992shape} which explore the intrinsic subspace structures. The geometric information present  in data  can be  exploited for various machine learning problems such as building effective classifiers~\cite{elhamifar2012block}, data modeling~\cite{turaga2008statistical} and  transformation learning~\cite{qiu2015learning}.


A notable work that uses  transformation learning to separate the subspaces is  by Qiu et al.~\cite{qiu2015learning}. In ~\cite{qiu2015learning},  a  linear transformation is learned  that preserves  the  low rank structure of  within class data  by minimizing the nuclear norm. Simultaneously, it also  maximizes  the separation  between the subspaces  of different class data by maximizing the  nuclear norm of the complete data matrix. The major disadvantage of this transformation is that  being a linear one, it cannot handle non-linearities present in data. We use the term non-linearity to capture any deviation from an ideal subspace structure.
Kernels can be used to overcome  the above mentioned drawback. But  kernels  do not preserve the subspace structure and  does not guarantee  separation of the subspaces.
These drawbacks of kernel  motivates us to use  neural networks for transformation learning for separating the  subspaces.

The use of autoencoder is quite popular in  various machine learning applications like image de-noising~\cite{vincent2010stacked},  dimensionality  reduction~\cite{hinton2006reducing}  and  data clustering~\cite{ji2017deep} to  name a few. Typically autoencoders  are used to generate data representations for above mentioned applications. In these cases training is carried out without using data labels.  But in our work, we train an autoencoder using  data labels for generating a representation to be used for classification. The cost function is chosen such that, in this process, the minimum principal angle is maximized.  
To the best of  our knowledge, this is the first  attempt  that uses an  autoencoder for  subspace specific  transformation learning for  classification.  We introduce  a  novel class-wise self expressiveness (CSSE)  layer in between encoder and decoder network  that preserves the  subspace structure in data in the feature space.   This  work is inspired by  subspace clustering work in~\cite{ji2017deep}  which  uses the self expressiveness layer along with an autoencoder. Our work is different in terms of the application, cost function and network architecture.  We would like to reiterate that this solution is superior only when  there are non-linearities present in data in the input space. In case of images, pixel space is termed as input space.
\section{The proposed approach}\label{sec2}
In this section, we  discuss the  self expressiveness property of  a linear subspace, proposed loss function,  network architecture and training strategy. 
\subsection{Self expressiveness within a    subspace}
Self expressiveness property states that  a data  point in a subspace can be expressed as  a  linear combination of other data points from the same subspace.  Let $\mathcal{Y}^i=\{\boldsymbol{y}_j\}_{j=1}^{n}$ denote a  set of $n$ data points drawn from a subspace  $\mathcal{S}_i \subset\mathbb{R}^D $.  Rewriting the set $\mathcal{Y}^i$ as matrix $Y^i \in \mathbb{R}^{D \times n}$, where each column represents a data point, self expressiveness can be represented  as
\begin{equation}
Y^i=Y^iW^i~~~ s.t. ~~~diag(W^i)=0,
\end{equation}

where,  $W^i \in \mathbb{R}^{n \times n}$ denotes the self representation coefficient matrix and the constraint on  $W^i$ avoids the trivial solution.  It may be noted that self expressiveness property for subspaces holds only for $n > d_s$, where  $d_s$  denotes  the dimension of the  subspace. Coefficient matrix $W^i$ can be obtained by solving:
\begin{equation}
\min_{W^i}  \parallel Y^i-Y^iW^i \parallel_F^2 + \lambda \parallel W^i \parallel_1~~~s.t.~~~diag(W^i)=0,
\end{equation}
where, $\parallel. \parallel_1$ denotes the $\ell_1$  norm of vectorized $W^i$ which promotes sparsity.

Since we are assuming  subspace structure for each class, a data point can be thought of as coming from union of subspaces. Let $S_1, \dots, S_c$ be the $c$ subspaces associated with $c$ classes. Self-expressiveness property is valid in each of the subspaces, \textit{i.e.},  a data point $\textbf{y}_i \in S_i$ is represented by points in $S_i$ and not by points in $S_j, j \neq i$ assuming non-overlapping subspaces.  Mathematically, this can be written as,
\begin{equation} \label{eq55}
Y =[ Y^1~~~ Y^2 ~\dots ~Y^c] \underbrace{\begin{bmatrix}
W^1 &0   &\dots &0\\
\vdots& &\ddots &\vdots\\
0 & 0 &\dots &W^c\\
\end{bmatrix} }_{W}~~~s.t.~~~diag(W)=0,
\end{equation}
where $Y= [Y^1,\dots,{Y}^c]$ denotes the $c$-class data drawn from a union of $c$ linear subspaces,  $\bigcup_{i=1}^c{\mathcal{S}_i}$, of $\mathbb{R}^D$, respectively.  $W$ is a block diagonal matrix where each block denotes the   representation  coefficients  corresponding to a particular class.  The optimization problem in (\ref{eq55})  ensures that   a data point of a class  should be represented only by   other data points of the  same  class. Representation coefficient can be obtained by solving,
\begin{equation}
\min_{W}  \sum_{i=1}^{c}||Y^i-Y^iW^i||_F^2 + \lambda||W||_1~~~s.t.~~~diag(W)=0.
\label{equ2}
\end{equation}

\begin{figure*}[]
	\centering
	\includegraphics[width=17cm, height=8cm]{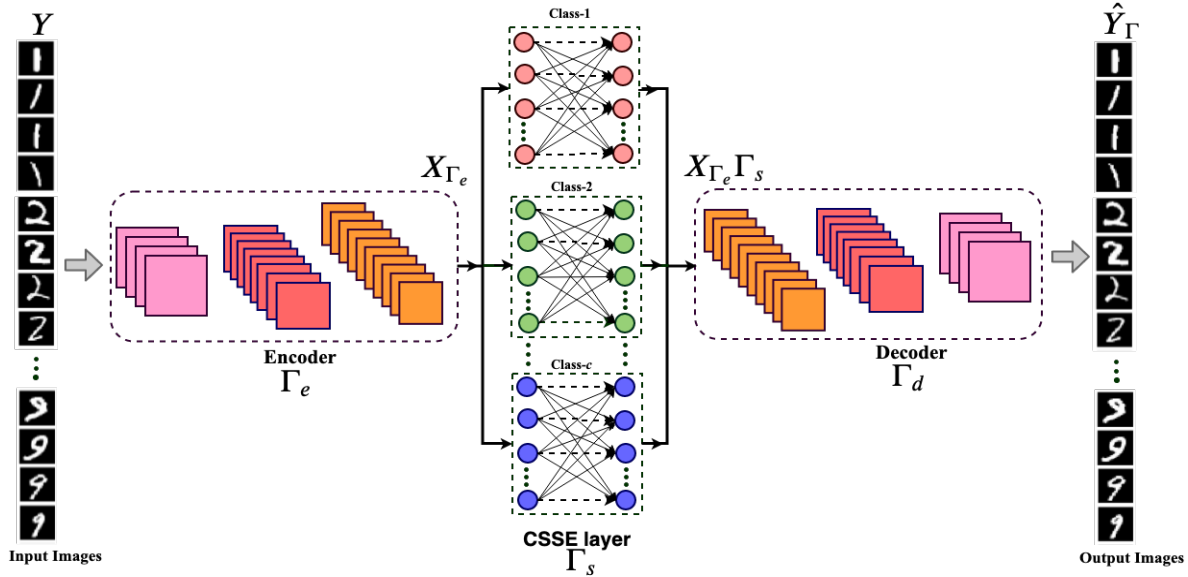}
	\caption{ The proposed network architecture  which consists of encoder network, novel CSSE layer and decoder network, respectively. }\label{fig1}
\end{figure*}
\subsection{Class specific  self expressiveness in convolutional   autoencoder} \label{sec2.2}
Typically autoencoders generate a shorter length representation of input data such that the reconstruction error is minimized. Our requirement  is a data representation which gives minimum reconstruction error while maintaining the subspace structure present in the data with increased separation between subspaces. The former issue is addressed in this subsection and the  latter in the following one.  Self expressiveness is built into the autoencoder by adding a block between encoder and decoder as shown in Fig. 1.  
The   convolutional autoencoder   is parametrized by $\Gamma =(\Gamma_e, ~\Gamma_s,~ \Gamma_d)$, where  $\Gamma_e$, $\Gamma_s$   and $\Gamma_d$ denote the weights of encoder, CSSE layer  and decoder networks,  respectively. $X_{\Gamma_e}$   denotes  the encoder output matrix corresponding to data matrix $Y=[Y^1, \dots, Y^c]  \in \mathbb{R}^{D \times N}$.  $X_{\Gamma_e}= [X_{\Gamma_e}^1, \dots, X_{\Gamma_e}^c ]$ with $X_{\Gamma_e}^i \in \mathbb{R}^{\hat{D} \times n_i}$  denoting the feature space representation for $ i^{th}$ class. The choice of $\hat{D}$ is explained in subsection \ref{sec:2.4}.  Incorporating the  notion of   class specific self expressiveness, the new cost  function for autoencoder becomes:
\begin{equation}
\begin{split}
& L(\Gamma_e, \Gamma_s, \Gamma_d) = \frac{1}{2}||Y -\hat{Y}_{\scriptsize{\Gamma}}||_F^2  +  \lambda_1\sum_{i=1}^{c}||X_{\Gamma_e}^i-X_{\Gamma_e}^i \Gamma_s^i||_F^2 + \lambda_2||\Gamma_s||_1 \quad s.t.\quad diag(\Gamma_s)=0,
\end{split}
\end{equation}
where, $\hat{Y}_\Gamma$ is the output of the decoder.
The  first term of the proposed  cost  function reduces MSRE (mean square  representation error); second and third terms together  preserves the subspace structure  in the feature space.  With this cost, the network behavior is equivalent to that of a kernel, the difference being the non-linear mapping is a learned one rather than  an analytic one.  Though this cost function preserves the subspace structure, it does not increase the distance between the subspaces. We  modify the cost function to equation (\ref{equ5})  to push the subspaces apart.

\subsection{Increasing the distance between subspaces } \label{sec2.3}
The  distance between two subspaces is measured in terms of principal angles~\cite{huang2015role}  defined as follows.
\begin{defn}
	Let $S_1, ~S_2 \subset \mathbb{R}^D$ be two $p$  dimensional linear subspaces. The principal angles  ($\theta_l \in [0 , \pi/2]; \hspace{5pt} l=1,...,p) $ between $S_1$ and $S_2$ , are
	defined recursively by
	
	\begin{equation*}
	\begin{split}
	\cos\theta_l= \underset{u_l \in S_1}{\max} \underset{v_l \in S_2}{\max} \hspace{5pt} u_l^Tv_l\\
	s.t.: \hspace{2pt}   \vert\vert u_l \vert\vert_2 = \vert\vert v_l \vert\vert_2  = 1 \\
	u_l^Tu_m = 0; \hspace{2pt} m=1,2,...,l-1\\
	v_l^Tv_m = 0; \hspace{2pt} m=1,2,...,l-1\\
	\end{split}
	\end{equation*}
	
\end{defn}

$\cos \theta_1$  is the cosine of  lowest principal angle which denotes the maximum correlation between two subspaces. Two subspaces are maximally separated if all  the principal angles  are 90 degrees, \textit{i.e.}, ${X_{\Gamma_e}^i}^\top{X_{\Gamma_e}^j} = \boldsymbol{0}$, where   $\boldsymbol{0} \in \mathbb{R}^{n_i \times n_j}$. It can be  achieved when $||{X_{\Gamma_e}^i}^\top X_{\Gamma_e}^j||_F^2$  is  minimized pairwise.  With the addition of this  term, new cost function becomes
\begin{equation}
\begin{split}
& L(\Gamma_e, \Gamma_s, \Gamma_d) =\frac{1}{2} ||Y -\hat{Y}_{\scriptsize{\Gamma}}||_F^2  + \lambda_1\sum_{i=1}^{c}||X_{\Gamma_e}^i-X_{\Gamma_e}^i \Gamma_s^i||_F^2 + \lambda_2||\Gamma_s||_1\\ &+ \lambda_3\sum_{i=1}^{c}\sum_{j=i+1}^{c}||{X_{\Gamma_e}^i}^\top X_{\Gamma_e}^j||_F^2 ~~~s.t.~~~diag(\Gamma_s)=0.
\end{split}\label{equ5}
\end{equation} 
\subsection{Network architecture and training strategy} \label{sec:2.4}
The  proposed network architecture is shown in Fig. \ref{fig1}.  The network  consists of three parts- encoder, CSSE layer and decoder. We use convolutional layers in both encoder and decoder networks  as it   requires less number of  parameters than fully connected layers and hence easy to train.   Kernels  in convolutional layers of both encoder and decoder use a stride of 2 in the both directions.  Note that in the CSSE layer, connections are preserved  between the data points of  same class only. There is no cross connection between the encoder outputs of different classes and hence preserves the within class subspace structure. The network acts as a transformation learner, \textit{i.e.}, it maps the different class data to the optimally separated subspaces using the proposed cost function  given in equation (\ref{equ5}). Training data is passed to the network  in a single batch  along with the class labels.  The explicit representations obtained from the encoder  are mapped to original input (pixel)  space using the decoder network.  Feature space dimension, $\hat{D}$, depends upon the  encoder design, \textit{i.e.}, number of conv layers, number of filters in each layer and stride of the filter. A desired feature space dimension can be obtained by adjusting the above mentioned parameters.

Let $t_j$ denote the number of filters in $j^{th}$ layer. $k_j \times k_j \times t_{j-1}$  is the size of filter in $j^{th}$ layer. $t_0$ is set to 1 as input to the network is gray scale images.
The total number of weight parameters  for  the  $j^{th}$ layer of encoder network are $k_j^2t_jt_{j-1}$.  Total weights for encoder network  with $m$  such layers are  $\sum_{j=1}^{m}k_j^2t_jt_{j-1}$. Bias parameters are excluded for the calculation.  Decoder being    similar in structure to encoder, cumulative sum of encoder and decoder weight parameters  is $2\sum_{j=1}^{m}k_j^2t_jt_{j-1}$. Weights in the CSSE layers depends on the  number of examples per class and total number of classes \textit{i.e.} $c(\sum_{i=1}^{c} n_i^2 -n_i)$.  Considering  an arbitrary example of 10 class data with 100 training images  per class,  a 3 layer network is chosen with  10, 20 and 30 filters per layer respectively with a kernel size of $5 \times 5 $. The total number of encoder and decoder parameters are $20250 \times 2$ and  CSSE layer parameters are approximately $10^5$. It is evident that network is mostly dominated by the parameters of CSSE layer.  With the increase in training data, number of parameters for CSSE also increases but remain fixed for encoder and decoder networks.  Though the number of parameters increase, most of the weights within a block of CSSE layer go to zero due to self expressiveness within a subspace and hence does not complicate the training procedure. 

The network is trained from scratch using Xavier initialization. This is a supervised learning framework as  the complete training data is passed in a single batch along with the class labels. The class labels ensures the  block diagonal connections in CSSE layer.  For loss minimization adaptive momentum (ADAM)  based gradient descent method is used. The method is a deterministic one rather than  stochastic as whole data is passed in a single batch.   Number of iterations  depends upon the type of  training data used. 

\section{Experimental analysis}\label{sec3}

\subsection{Dataset Used}
\vspace{-1mm}
\subsubsection{Extended Yale  b face~\cite{1407873}}  The dataset consists of the frontal face images of 38 individual subjects. For each subject, images are captured under 64  different illumination  conditions  having a total of 2414  gray-scale images. In our experiment, we resize  all the images to a fixed  size of $48 \times 42$.  Half of the images are  selected randomly for training and remaining half for testing.
\subsubsection{MNIST} MNIST is a handwritten digit dataset used for digit recognition task. It consists of 60000 training images and 10000 testing images  of size  $28 \times 28$. We randomly select  only  100 images per digit for training  with a total of 1000 images for 10  classes. 
\subsubsection{Coil20~\cite{nene1996columbia}} This is an object dataset consisting of 20 different object categories.  72 images per object  are captured at different viewing angles  with a total of 1440 images.   We resize all the images to a size of $32 \times 32$. The dataset is divided into  two equal parts for training and testing.  We adapt two modes of settings. In setting 1, 36 images  per object are randomly selected  for training while in setting 2, consecutive 36 images are selected. Setting 2 is more challenging as images kept in training and testing are captured at totally different viewing angles. 

\subsection{Parameter settings}
For Extended Yale b dataset, we empirically select a 3-layer encoder-decoder network.  First layer has 30 channels with kernel size of $5 \times 5$. Second and third layers  have 25 and 20 number of channels,  respectively with kernel size of $3 \times 3$.  Since face images exhibit a  well defined subspace structure in input space, no ReLU is added. The   non-linearity in network is due to biases in filters only.   This network architecture leads to the generation of a 720 dimensional feature vector representation corresponding to each input image. $\lambda_1, ~\lambda_2 ~\text{and}~ \lambda_3$ are  empirically chosen as 3, 1 and 100, respectively.  Choice of $\lambda 's$ is empirical. Data is passed in a single batch of size 1216 along with class labels. The learning rate is set to $10^{-3}$.

For MNIST dataset,  a 3-layer encoder-decoder network is used with 30, 20 and 10 number of channels in each layer, respectively.  Kernel size is kept same as in previous case. Except for the addition of ReLU to the second layer of both encoder and decoder, the  architecture  is same  as in previous case.  This ReLU handles the non-linearities of input space. This network architecture leads to the generation of a 160 dimensional feature vector representation corresponding to each input image.  $\lambda_1, ~\lambda_2$ and $ \lambda_3$ are empirically  chosen as 0.5, 0.1 and 50, respectively. 

For Coil20 dataset, a   similar architecture as that of MNIST is used  which generates a 160 length representation in feature space.  $\lambda_1,~ \lambda_2$ and  $ \lambda_3$ are empirically chosen as 1, 1 and 70, respectively.  
\vspace{-3mm}
\subsection{Classification performance}
The proposed approach is compared with  a $k$-NN classifier applied on pixel features  and  two linear transformation learning approaches: orthogonalizing transformation (OT) ~\cite{qiu2015learning}  and low rank transformation (LRT) ~\cite{qiu2015learning}.  $k$-NN and sparse representation (SR)~\cite{wright2008robust} based classifier are used over the obtained feature space representations. Classification accuracy  for an average of  4 random trials is reported  in Table \ref{Tab2} for all the datasets. A comparable classification performance can be seen for Extended Yale face dataset. The reason is that the  face images are captured in varying illumination keeping  pose and expression fixed.  This results in a better subspace structure in input space leading to almost no non-linearities and hence  other transformations also work well. A better classification performance for Mnist digit dataset is also observed by considering only a small set of training samples. Experimentation performed on Coil20 dataset in setting 1 shows a comparable performance for all. The reason being that the  36 data points for each class are sampled randomly for both training and testing, respectively. This leads to the presence of data points in nearby viewing angles in both the training and testing  and hence gives a good performance. However this is not the case under setting 2 where  the 36 consecutive viewing angles are kept for training and remaining for testing.  For this, a significant improvement in performance is seen.

\begin{table*}[h] 
	\centering
	\caption{Comparison of classification accuracies  with state-of-the-art methods. }
	\label{Tab2}
	
	{
		\begin{tabular}{lcccc}
			\hline
			\textbf{Methods}                                                                         & \textbf{Extended Yale b face }   & \textit{\textbf{Mnist digit}}   & \textbf{Coil20 (setting1)}          & \textbf{Coil20(setting2)}       \\ \hline \hline
			k- Nearest Neighbor   & 76.34  $\pm$  0.29                                   & 88.02 $\pm$ 0.67                                  & 98.64 $\pm$ 0.07                                    &  81.99 $\pm$ 3.02                                                                 \\ \hline
			
			Orthogonalizing transformation                &  93.95 $\pm$  0.91                                  &  82.80 $\pm$  1.13                               &  {97.71 $\pm$  0.68  }                           &   82.64 $\pm$  2.61                                        
			\\ \hline
			Low rank transformation             &  93.31 $\pm$  1.03                                 &  86.27 $\pm$ 0.22                               &   98.80 $\pm$  0.23                            &   83.57 $\pm$  5.04                                      
			\\ \hline
			The proposed approach + $k$ - NN            &  {91.10 $\pm$  0.71  }                              & { 88.53 $\pm$  0.25 }                                &    {98.42 $\pm$  0.47 }                            &   { 85.31 $\pm$  1.75  }                                   
			\\ \hline
			The proposed approach + SR           &  \textbf{94.57 $ \pm$  0.38  }                              & \textbf{ 91.35 $\pm$  0.35}                                &    \textbf{98.80$ \pm$  0.28 }                            &   \textbf{ 86.74 $\pm$  1.39  }                                   
			\\ \hline
			
		\end{tabular}
	}
\end{table*}

\section{Analysis}\label{sec4}
\subsection{Role of CSSE layer}  
\begin{figure}[!htb]
	\centering
	\begin{subfigure}[b]{0.15\textwidth}
		\includegraphics[width=\linewidth]{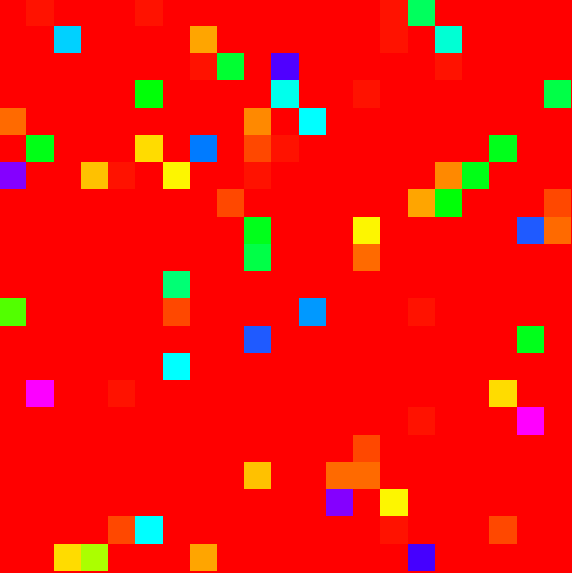}
		\caption{}
	\end{subfigure}%
	\hspace{5mm}
	\begin{subfigure}[b]{0.15\textwidth}
		\includegraphics[width=\linewidth]{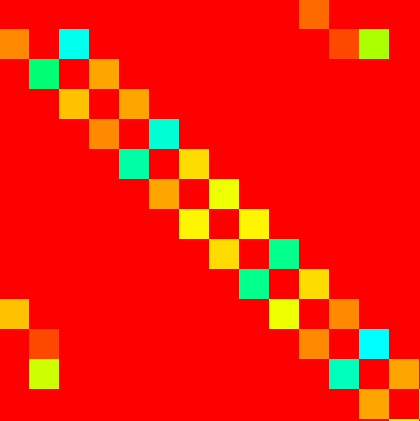}
		\caption{}
	\end{subfigure}%
	\caption{Weight parameters for a class specific block of  CSSE layer for Coil20 dataset in setting 1 and 2 respectively.}\label{weight}
\end{figure}

CSSE layer is used to preserve  the subspace structure from the input space  to the  feature space. 
Fig. \ref{weight} shows the weight parameters for CSSE layer for Coil20 dataset obtained after training. The weight matrices are  shown for (a) randomly sampled training data (setting1) and (b) consecutive sampled training data (setting 2). A sparsity pattern within a class specific  block can be observed in both the settings but nature is different. In Fig. \ref{weight}(a)  a random sparsity  pattern is observed while   in Fig. \ref{weight}(b)  non-zero weights are observed along the diagonal entries. The reason is quite intuitive as the training data is structured in the $2^{nd}$ case,  all the training data points are arranged in slightly varying poses.  Hence a data point can be best represented only by its nearby poses that leads to non-zero  entries only along the diagonal values.  

\vspace{-3mm}
\subsection{Separation between subspaces}
The cosine of lowest principle angle between two subspaces can be computed as,
$\cos \theta = \max_{u \in \mathcal{S}_i} \max_{v \in \mathcal{S}_j} \langle u, v\rangle,$
where  $\mathcal{S}_i$ denotes the subspace for $i^{th}$ class data. Table \ref{angle} shows the angle (in degree)  between the class specific subspaces for few  selected objects from Coil20 dataset.  We model each class data as a  seven dimensional subspace (decided by using singular values). The subspace  bases  are  generated using singular value decomposition of class specific data matrix.  We can observe that there is a significant increase in the lowest principal angle which leads to better separability of data in the feature space. 
\begin{table*}[h]
	\caption{\small{Table shows the angle (in degree)  between the class specific subspaces for few  selected objects from Coil20 dataset. [Left]  lowest principal subspace angle in input(pixel)  space [Right] lowest principal subspace angle in feature (encoder output) space.} }
	\centering
	
	\begin{tabular}{|l|c|c|c|c|c|}
		\hline
		\textbf{\begin{tabular}[c]{@{}l@{}}object\\  index\end{tabular}} & \multicolumn{1}{l|}{\textbf{3}} & \multicolumn{1}{l|}{\textbf{5}} & \multicolumn{1}{l|}{\textbf{6}} & \multicolumn{1}{l|}{\textbf{9}} & \multicolumn{1}{l|}{\textbf{19}}  \\ \hline
		\textbf{3}                                                       & 0          & 19.72      & 20.04      & 27.48      & 15.89       \\  \hline 
		\textbf{5}                                                       & 19.72     & 0          & 21.41     & 15.96      & 10.03       \\ \hline
		\textbf{6}                                                       & 20.04      & 21.41      & 0          & 27.18      & 16.32      \\ \hline
		\textbf{9}                                                       & 27.48      & 15.96      & 27.18      & 0          & 15.30      \\  \hline
		\textbf{19}                                                      & 15.89      & 10.03      & 16.32     & 15.30      & 0           \\ \hline
	\end{tabular}
	\hspace{1mm}
	\begin{tabular}{|l|c|c|c|c|c|}
		\hline
		\textbf{\begin{tabular}[c]{@{}l@{}}object\\  index\end{tabular}} & \multicolumn{1}{l|}{\textbf{3}} & \multicolumn{1}{l|}{\textbf{5}} & \multicolumn{1}{l|}{\textbf{6}} & \multicolumn{1}{l|}{\textbf{9}} & \multicolumn{1}{l|}{\textbf{19}} \\ \hline
		\textbf{3}                                                       & 0                               & 50.55                          & 35.01                           & 54.21                          & 24.39                           \\ \hline 
		\textbf{5}                                                       & 50.55                         & 0                               & 55.40                          & 51.01                           & 52.58                            \\ \hline
		\textbf{6}                                                       & 35.01                         & 55.40                         & 0                               & 56.37                          & 29.92                           \\ \hline
		\textbf{9}                                                       & 54.21                          & 51.01                           & 56.37                          & 0                               & 51.08                           \\ \hline
		\textbf{19}                                                      & 24.39                        & 52.58                           & 29.92                           & 51.08                          & 0                                \\ \hline
	\end{tabular}\label{angle}
\end{table*}

%
%
%
%

\section{Conclusion} \label{sec5} 
In this work, we have proposed a novel  autoencoder based  architecture which does transformation learning for classification. The introduction of new  CSSE layers helps in preserving the subspace structure in the feature space due to the  self expressiveness.  The class specific subspaces are separated optimally   by using  the proposed cost  function. It is also evident by looking at the smallest principal subspace angles in feature space. We have shown that the proposed approach works significantly better than the state-of-the-art methods.

\bibliographystyle{IEEEbib}
\bibliography{krishan_work.bib}

\begin{thebibliography}{10}

\bibitem{basri2003lambertian}
Ronen Basri and David~W Jacobs,
\newblock ``Lambertian reflectance and linear subspaces,''
\newblock {\em IEEE Transactions on Pattern Analysis \& Machine Intelligence},
  vol. 25, no. 2, pp. 218--233, 2003.

\bibitem{hastie1998metrics}
Trevor Hastie and Patrice~Y Simard,
\newblock ``Metrics and models for handwritten character recognition,''
\newblock {\em Statistical Science}, pp. 54--65, 1998.

\bibitem{murase1995visual}
Hiroshi Murase and Shree~K Nayar,
\newblock ``Visual learning and recognition of 3-{D} objects from appearance,''
\newblock {\em International {J}ournal of {C}omputer {V}ision}, vol. 14, no. 1,
  pp. 5--24, 1995.

\bibitem{tomasi1992shape}
Carlo Tomasi and Takeo Kanade,
\newblock ``Shape and motion from image streams under orthography: a
  factorization method,''
\newblock {\em International {J}ournal of {C}omputer {V}ision}, vol. 9, no. 2,
  pp. 137--154, 1992.

\bibitem{elhamifar2012block}
Ehsan Elhamifar and Ren{\'e} Vidal,
\newblock ``Block-sparse recovery via convex optimization,''
\newblock {\em IEEE Transactions on Signal Processing}, vol. 60, no. 8, pp.
  4094--4107, 2012.

\bibitem{turaga2008statistical}
Pavan Turaga, Ashok Veeraraghavan, and Rama Chellappa,
\newblock ``Statistical analysis on {S}tiefel and {G}rassmann manifolds with
  applications in computer vision,''
\newblock in {\em IEEE Conference on Computer Vision and Pattern Recognition
  (CVPR)}. IEEE, 2008, pp. 1--8.

\bibitem{qiu2015learning}
Qiang Qiu and Guillermo Sapiro,
\newblock ``Learning transformations for clustering and classification,''
\newblock {\em The Journal of Machine Learning Research}, vol. 16, no. 1, pp.
  187--225, 2015.

\bibitem{vincent2010stacked}
Pascal Vincent, Hugo Larochelle, Isabelle Lajoie, Yoshua Bengio, and
  Pierre-Antoine Manzagol,
\newblock ``Stacked denoising autoencoders: Learning useful representations in
  a deep network with a local denoising criterion,''
\newblock {\em Journal of Machine Learning Research}, vol. 11, no. Dec, pp.
  3371--3408, 2010.

\bibitem{hinton2006reducing}
Geoffrey~E Hinton and Ruslan~R Salakhutdinov,
\newblock ``Reducing the dimensionality of data with neural networks,''
\newblock {\em Science}, vol. 313, no. 5786, pp. 504--507, 2006.

\bibitem{ji2017deep}
Pan Ji, Tong Zhang, Hongdong Li, Mathieu Salzmann, and Ian Reid,
\newblock ``Deep subspace clustering networks,''
\newblock in {\em Advances in Neural Information Processing Systems}, 2017, pp.
  24--33.

\bibitem{huang2015role}
Jiaji Huang, Qiang Qiu, and Robert Calderbank,
\newblock ``The role of principal angles in subspace classification,''
\newblock {\em IEEE Transactions on Signal Processing}, vol. 64, no. 8, pp.
  1933--1945, 2015.

\bibitem{1407873}
{Kuang-Chih Lee}, J.~{Ho}, and D.~J. {Kriegman},
\newblock ``Acquiring linear subspaces for face recognition under variable
  lighting,''
\newblock {\em IEEE Transactions on Pattern Analysis and Machine Intelligence},
  vol. 27, no. 5, pp. 684--698, May 2005.

\bibitem{nene1996columbia}
Sameer~A Nene, Shree~K Nayar, Hiroshi Murase, et~al.,
\newblock ``Columbia object image library (coil-20),''
\newblock 1996.

\bibitem{wright2008robust}
John Wright, Allen~Y Yang, Arvind Ganesh, S~Shankar Sastry, and Yi~Ma,
\newblock ``Robust face recognition via sparse representation,''
\newblock {\em IEEE {T}ransactions on {P}attern {A}nalysis and {M}achine
  {I}ntelligence}, vol. 31, no. 2, pp. 210--227, 2008.

\end{thebibliography}

\end{document}